\documentclass[10pt,twocolumn,letterpaper]{article}

\usepackage[pagenumbers]{cvpr} 

\usepackage{graphicx}
\usepackage{amsmath}
\usepackage{amssymb}
\usepackage{booktabs}
\usepackage{balance}
\usepackage{lipsum}
\usepackage{caption}
\usepackage{algorithm}
\usepackage{algorithmic}
\usepackage{enumerate}
\usepackage{amsmath,amsthm,amssymb}
\usepackage{engord}
\usepackage{bm}
\usepackage{bbm}
\usepackage{amstext}
\usepackage{comment}
\usepackage{color}
\usepackage[x11names,table, dvipsnames]{xcolor}
\usepackage{ctable}
\usepackage{booktabs}
\usepackage{multirow}
\usepackage{mathtools}
\usepackage{setspace}
\usepackage{pifont}
\usepackage{dsfont} 
\usepackage{cancel}
\usepackage[toc,page]{appendix}

\usepackage[pagebackref,breaklinks,colorlinks]{hyperref}
\usepackage[capitalize]{cleveref}
\usepackage{hhline}

\definecolor{Gray}{gray}{0.9}
\definecolor{White}{gray}{1}
\definecolor{WhiteGray}{rgb}{0.9, 0.9, 0.9}
\definecolor{DGray}{gray}{0.8}
\definecolor{DDDDGray}{gray}{0.3}
\definecolor{citecolor}{HTML}{0071bc}

\definecolor{DeltaColor}{rgb}{0.039,0.73,0.71}
\definecolor{SigmaColor}{rgb}{0.98,0.45,0.0}
\definecolor{AlphaColor}{rgb}{0,0,0.8}
\definecolor{BetaColor}{rgb}{0.8,0,0.8}
\definecolor{GammaColor}{rgb}{0.514,0.34,0.224}
\definecolor{EpsilonColor}{rgb}{0.353,0.725,0.906}
\definecolor{GreenColor}{rgb}{0.137,0.573,0.565}
\definecolor{RedColor}{rgb}{0.949,0.275, 0.224}

\DeclareMathAlphabet\mathbfcal{OMS}{cmsy}{b}{n}

\DeclareMathOperator*{\argmin}{arg\,min}

\newcommand{\colorRef}[1]{\textcolor{red}{#1}}
\crefname{figure}{\colorRef{Fig.}}{\colorRef{Figs.}}
\Crefname{figure}{\colorRef{Figure}}{\colorRef{Figures}}
\crefname{section}{\colorRef{Sec.}}{\colorRef{Secs.}}
\Crefname{section}{\colorRef{Section}}{\colorRef{Sections}}

\crefname{table}{\colorRef{Tab.}}{\colorRef{Tabs.}}
\Crefname{table}{\colorRef{Table}}{\colorRef{Tables}}
\Crefname{equation}{\colorRef{Eq.}}{\colorRef{Eqs.}}
\Crefname{equation}{\colorRef{Equation}}{\colorRef{Equation}}

\newtheorem{property}{Property}

\newcommand\method{Color-NeuS\xspace}

\newcommand{\projectURL}{\href{https://colmar-zlicheng.github.io/color_neus}{\tt{colmar-zlicheng.github.io/color\_neus}}}

\newcommand{\qheading}[1]{\vspace{5pt}\noindent\mbox{\textbf{#1}\;}}

\newcommand{\eofparagraph}{\hfill \scalebox{0.5}{$\blacksquare$}}

\newcommand{\del}[1]{}

\newcommand{\emath}{\ensuremath}

\newcommand{\customfootnotetext}[2]{{
      \renewcommand{\thefootnote}{#1}
      \footnotetext[0]{#2}}}

\newlength\savewidth\newcommand\shline{\noalign{\global\savewidth\arrayrulewidth
    \global\arrayrulewidth 1pt}\hline\noalign{\global\arrayrulewidth\savewidth}}

\def\paperID{27} 
\def\confName{3DV\xspace}
\def\confYear{2024\xspace}

\title{Color-NeuS: Reconstructing Neural Implicit Surfaces with Color}

\author{
{Licheng Zhong\textsuperscript{1~$\star$}\;
                Lixin Yang\textsuperscript{1,2~$\star$}\;
                Kailin Li\textsuperscript{1}\;
                Haoyu Zhen\textsuperscript{1}\;
                Mei Han\textsuperscript{3}\;
                Cewu Lu\textsuperscript{1,2~$\boldsymbol{\dagger}$}}\\
{
\small
{$^{1}$Shanghai Jiao Tong University}\quad\quad
{$^{2}$Shanghai Qi Zhi Institute}\quad\quad
{$^{3}$National University of Singapore}\quad\quad
}\\
{
\tt\small  \{{zlicheng}, {siriusyang}, {kailinli}, {anye\_zhen}, {lucewu}\}@{sjtu.edu.cn}
}
{
\tt\small  \{{hanmei}\}@{u.nus.edu}
} \\
}

\begin{document}
\maketitle
\customfootnotetext{$\star$}{
    These authors contributed equally.
}
\customfootnotetext{$\dagger$}{
    Cewu Lu is the corresponding author. He is the member of Qing Yuan Research Institute and MoE Key Lab of Artificial Intelligence, AI Institute, Shanghai Jiao Tong University, and Shanghai Qi Zhi Institute, China.
}

\begin{abstract}
    The reconstruction of object surfaces from multi-view images or monocular video is a fundamental issue in computer vision. However, much of the recent research concentrates on reconstructing geometry through implicit or explicit methods. In this paper, we shift our focus towards reconstructing mesh in conjunction with color. We remove the view-dependent color from neural volume rendering while retaining volume rendering performance through a relighting network. Mesh is extracted from the signed distance function (SDF) network for the surface, and color for each surface vertex is drawn from the global color network. To evaluate our approach, we conceived a in hand object scanning task featuring numerous occlusions and dramatic shifts in lighting conditions. We've gathered several videos for this task, and the results surpass those of any existing methods capable of reconstructing mesh alongside color. Additionally, our method's performance was assessed using public datasets, including DTU, BlendedMVS, and OmniObject3D. The results indicated that our method performs well across all these datasets. Project page: \projectURL.
\end{abstract}
\vspace{-10pt}
\section{Introduction}
\label{sec:intro}
\vspace{-5pt}
The endeavor of reconstructing 3D objects from 2D images is a pivotal and ongoing challenge in the domains of computer vision and graphics. Previously, the structure-from-motion (SFM) method \cite{schoenberger2016sfm, schoenberger2016mvs} was widely used to reconstruct 3D objects from 2D images. However, it often struggled with scenes lacking texture or those with repetitive patterns, leading to ambiguous correspondences and incorrect depth estimations. Another limitation of SFM was its inability to effectively handle occlusions. When an object in the scene was partially obscured, it would often lead to errors in reconstruction. A further limitation of SFM lay in its dependency on point cloud representation, which falls short of accomplishing a fully dense reconstruction.
Recently, the landscape of this field has been evolving, with a burgeoning interest in the investigation of implicit neural surfaces via volume rendering \cite{wang2021neus} which can represent pixel-level fine surface. It's based on the neural radiance field (NeRF) \cite{mildenhall2020nerf}.

\begin{figure}[!t]
    \begin{center}
        \includegraphics[width=\linewidth]{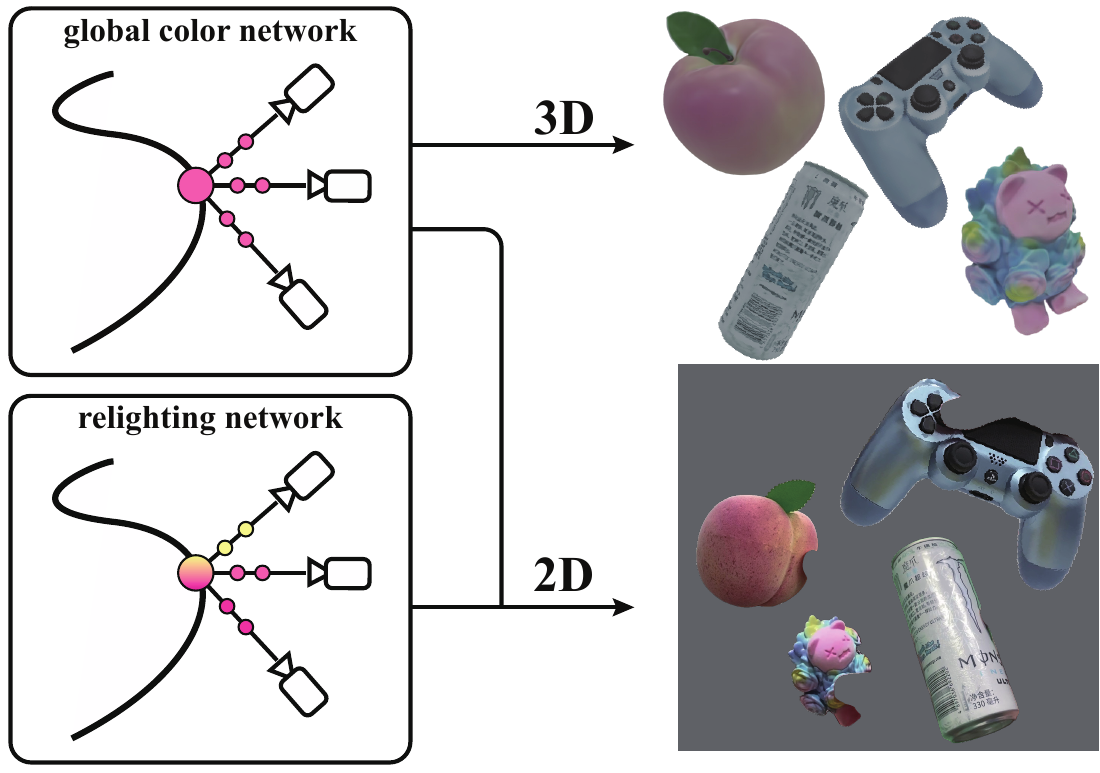}
    \end{center}
    \vspace{-15pt}
    \caption{\method decouple the view-dependent component from the color network and employs a relighting network to account for the view-dependent aspect. This allows \method to reconstruct a 3D implicit surface with color, while maintaining its ability for 2D volume rendering.}
    \label{fig:teaser}
    \vspace{-10pt}
\end{figure}

Pioneering works like NeRF \cite{mildenhall2020nerf} and its successors \cite{yu2020pixelnerf, mvsnerf, barron2021mipnerf, liu2022neuray, xu2022point} have convincingly demonstrated the power of neural networks in representing continuous 3D scenes. This is achieved by learning a mapping from 3D coordinates to volume density and view-dependent color, leading to highly effective novel view synthesis. Subsequently, NeuS \cite{wang2021neus} extended this concept, leveraging signed distance functions (SDF) to reconstruct a neural implicit surface. However, NeuS's scope remains limited to the reconstruction of a mesh without vertex color.
This limitation arises as the color of each point in NeuS's neural volume rendering is determined by both its position and viewing ray direction. So, in the context of mesh reconstruction, where the view direction is absent, NeuS cannot assigns a specific color value to each point.
In addition, most of the NeRF relighting works focus on 2D image rendering rather than textured surfaces reconstruction.
To address these limitation, in this paper, we present a novel method for the reconstruction of neural implicit surfaces incorporating \textbf{view-independent} color.

We propose \method, a \textbf{NeuS}-compatible approach that facilitates the reconstruction of 3D surface and global vertex \textbf{Color}s, independent of the viewpoint. Simultaneously, it upholds NeuS's robust functionalities in rendering 2D images and executing 3D surface reconstruction.
To this end, we replace the singular view-dependent color component in the neural volume rendering with the integration of a view-independent global color variable and a view-dependent relighting effect as shown in \cref{fig:teaser}.
This not only enables our model to be trained for standard volume rendering but also paves the way for the learning of global vertex colors.
Importantly, \method naturally handles substantial reflection on the object's surface, and can deal with intermittent occlusion during dynamic interaction between objects (see \cref{fig:occlu_reflect}).

The effectiveness of our \method has been extensively validated on various datasets including the DTU \cite{jensen2014large} , BlendedMVS \cite{yao2020blendedmvs}, and OmniObject3D \cite{wu2023omniobject3d} datasets.
To demonstrate the superiority of our method, we conducted comparative evaluations with laser scanning and well-established methods such as COLMAP \cite{schoenberger2016sfm, schoenberger2016mvs} and HHOR \cite{huang2022hhor}.
The results reveal that \method successfully reconstructs the object's surface while extracting a reasonable texture.
To underscore its practical application,
we apply \method to a real-world challenging  task: hand-held object scanning \cite{huang2022hhor, hampali2023inhandscan}. As part of this process, we collect a dataset for validation, which includes 3D scans of objects, videos of moving objects held in hand, and object-centric camera transformations.

Our contributions can be summarized as follows:

\begin{itemize}
    \item We propose a novel method for the reconstruction of a neural implicit surface with color that can be applied to any NeuS-like models.
    \item We decouple the view-dependent process in neural volume rendering, enabling the handling of occlusion and reflection while obtaining global vertex colors.
    \item We devise a challenging task of in-hand object scanning for the reconstruction of mesh with color and compile a real-world dataset for this task.
\end{itemize}
\section{Related Work}
\label{sec:formatting}
\vspace{-5pt}

\qheading{Neural Radiance Field.} Following the pioneering work of NeRF \cite{mildenhall2020nerf}, a variety of studies have emerged in the realm of neural implicit fields. Notable among them, such as NeRF$--$ \cite{wang2021nerfmm}, NoPe-NeRF \cite{bian2022nopenerf}, BARF \cite{lin2021barf}, and GNeRF \cite{meng2021gnerf}, have been investigating methods to estimate camera pose while training the implicit field. Concurrently, efforts such as Pixel-NeRF \cite{yu2020pixelnerf}, MVSNeRF \cite{mvsnerf}, and IBRNet \cite{wang2021ibrnet} are focusing on the generalization of neural radiance fields. In parallel, research projects like KiloNeuS \cite{esposito2022kiloneus}, NeX \cite{Wizadwongsa2021NeX}, NeRV \cite{nerv2021}, NeRD \cite{boss2021nerd}, PhySG \cite{physg2021}, and NeRFactor \cite{nerfactor2021} have been focusing on illumination and relighting in neural fields. Our method is compatible with neural volume rendering and retents the capacity of new view synthesis.

\qheading{Surface Reconstruction.} Implicit Differentiable Renderer (IDR) \cite{yariv2020multiview} represents geometry as the zero level set from Signed Distance Function (SDF) by leveraging implicit geometric regularization (IGR) \cite{eikonal}. NeuS \cite{wang2021neus} amalgamates the SDF field and volume rendering to reconstruct surfaces. Both VolSDF \cite{yariv2021volume} and UNISURF \cite{Oechsle2021ICCV} combine implicit surface representation with volume rendering: VolSDF \cite{yariv2021volume} disentangles appearance from geometry, while UNISURF \cite{Oechsle2021ICCV} formulates implicit surface models and radiance fields in a cohesive manner. PET-NeuS \cite{wang2023petneus}, an extension of NeuS \cite{wang2021neus}, introduces new components such as an unique positional encoding type, tri-plane representation, and learnable convolution operations. However, these methodologies do not factor in the global view-independent color of the surface. Our method focuses on the extraction of surface color from a global color network while maintaining geometry learning and image rendering from a relighting network.

\qheading{In-hand Object Scanning.} ObMan \cite{hasson19_obman} presents an end-to-end learnable model that employs a unique contact loss, which encourages physically plausible hand-object configurations. For the in-hand object scanning task, IHOI \cite{ye2022hand} reconstructs from a single RGB image, taking advantage of the estimated hand pose. BundleSDF \cite{wen2023bundlesdf}, on the other hand, estimates the object pose using RGBD input sequential images, all the while reconstructing the implicit surface represented by the Signed Distance Function (SDF). HHOR \cite{huang2022hhor} also utilizes SDF to represent the object surface, but it reconstructs the object in tandem with the hand, where the object is firmly held. A recent work \cite{hampali2023inhand} reconstructs the object from an RGB sequence and estimates the camera pose simultaneously by using an occupancy field to represent the surface. However, it makes the assumption that the light source is distant and the direction of light remains unchanged. In contrast, our method can handle arbitrary lighting conditions to model the object appearance.

\section{Preliminary}

We first introduce Neural Implicit Surface (NeuS) \cite{wang2021neus}, which is the basis of our method. Given a camera with known intrinsic parameters, we can represent a ray in the camera coordinate system as
\begin{equation}
    \bm{p}(z)=\bm{o}+z\bm{d},
    \label{eq:ray}
\end{equation}
where $\bm{o}$ and $\bm{d}$ are the origin and the direction of the ray, respectively, and $z$ is the distance from the origin to the point on the ray.
Then, a MLP network $\mathcal{G}$ is used to encode $\bm{p}$ to its signed distance function (SDF) $s(\bm{p})$ and feature vector $f(\bm{p})$, as
\begin{equation}
    [s(\bm{p}), f(\bm{p})]=\mathcal{G}(\bm{p}).
    \label{eq:sdf_net}
\end{equation}

With the position $\bm{p}$, direction $\bm{d}$, feature vector $f(\bm{p})$ and gradient $g(\bm{p})$ as input, another MLP $\mathcal{M}$ outputs the color of the query point, as
\begin{equation}
    c(\bm{p}, \bm{d})=\mathcal{M}(\bm{p}, \bm{d}, f(\bm{p}), g(\bm{p})),
    \label{eq:color_net}
\end{equation}
where $g(\bm{p})=\nabla s(\bm{p})$ is the gradient of SDF at point $\bm{p}$.
Finally, the color of the query pixel $C$ is obtained by integrating the color along the ray,
\begin{equation}
    C = \int_{z_n}^{z_f}w(z)c(\bm{p}, \bm{d})dz,
    \label{eq:volume_render}
\end{equation}
\begin{equation}
    w(z)=\exp\big(-\int_{z_n}^{z}\sigma(t)dt \big) * \sigma(z),
    \label{eq:weights}
\end{equation}
\begin{equation}
    \sigma(\bm{p})=\frac{\alpha e^{-\alpha s(\bm{p})}}{(1+e^{-\alpha s(\bm{p})})^2},
    \label{eq:sigma}
\end{equation}
where $\sigma(\bm{p})$ designates the density of $\bm{p}$ with $\alpha \in \mathbb{R}^1$ acting as a learnable parameter. The term $w(z)$ refers to the weight assigned to the color at point $z$. Additionally, $z_n$ and $z_f$ represent the near and far plane of the camera, respectively.

Based on the input \emath{\bm{d}} in \cref{eq:volume_render}, the output per-vertex color is \textbf{view-dependent}.
Consequently, the original NeuS avoids incorporating color, focusing solely on the reconstruction of surface shape.

\section{Method}
\label{sec:method}

Our goal is to both extract the color and geometry of objects,
\begin{equation}
    c_g(\boldsymbol{x}), \boldsymbol{x} \in \mathcal{S},
\end{equation}
where $\mathcal{S}=\left\{ \boldsymbol{x} \in \mathbb{R}^3|s(\boldsymbol{x}) = 0\right\}$ represents the object surface (a collection of mesh vertices) that is characterized by a set of points with zero-level signed distance value.

\begin{property}
    (\cite[Sec.3.1]{wang2021neus})
    NeuS possesses an advantageous property  wherein the standard deviation of the density $\sigma(\boldsymbol{p})$ is dictated by the trainable parameter $1/\alpha$, which progressively approaches zero as the network training reaches convergence.
    \eofparagraph
    \label{property:neus}
\end{property}
\noindent Implied by \cref{property:neus}, the density $\sigma(\bm{p})$ of the neural radiance field substantially condenses on the surface $\mathcal{S}$.
This provides a focal point for us to extract surface (per-vertex) color of the object, when the networking training reaches convergence.

\subsection{Naive Solution}
\label{sec:naive}
One solution that might seem intuitive is to simply expunge the view-dependent term in \cref{eq:color_net}, as:
\begin{equation}
    c(\boldsymbol{p})=\mathcal{M}(\boldsymbol{p}, f(\boldsymbol{p}), g(\boldsymbol{p})).
    \label{eq:naive}
\end{equation}

Regrettably, undertaking this procedure can potentially impair the learned geometry and appearances within the neural radiance field.
This is attributed to the incapacity of neural radiance field to accurately express light variations of points across different directions, when a view-dependent term is absent. Furthermore, such an approach may also culminate in the breakdown of the SDF field, subsequently triggering fragmentation of the surface.
The qualitative results of naive solution are displayed in \cref{fig:exp_iho}.

\begin{figure}[!t]
    \begin{center}
        \includegraphics[width=\linewidth]{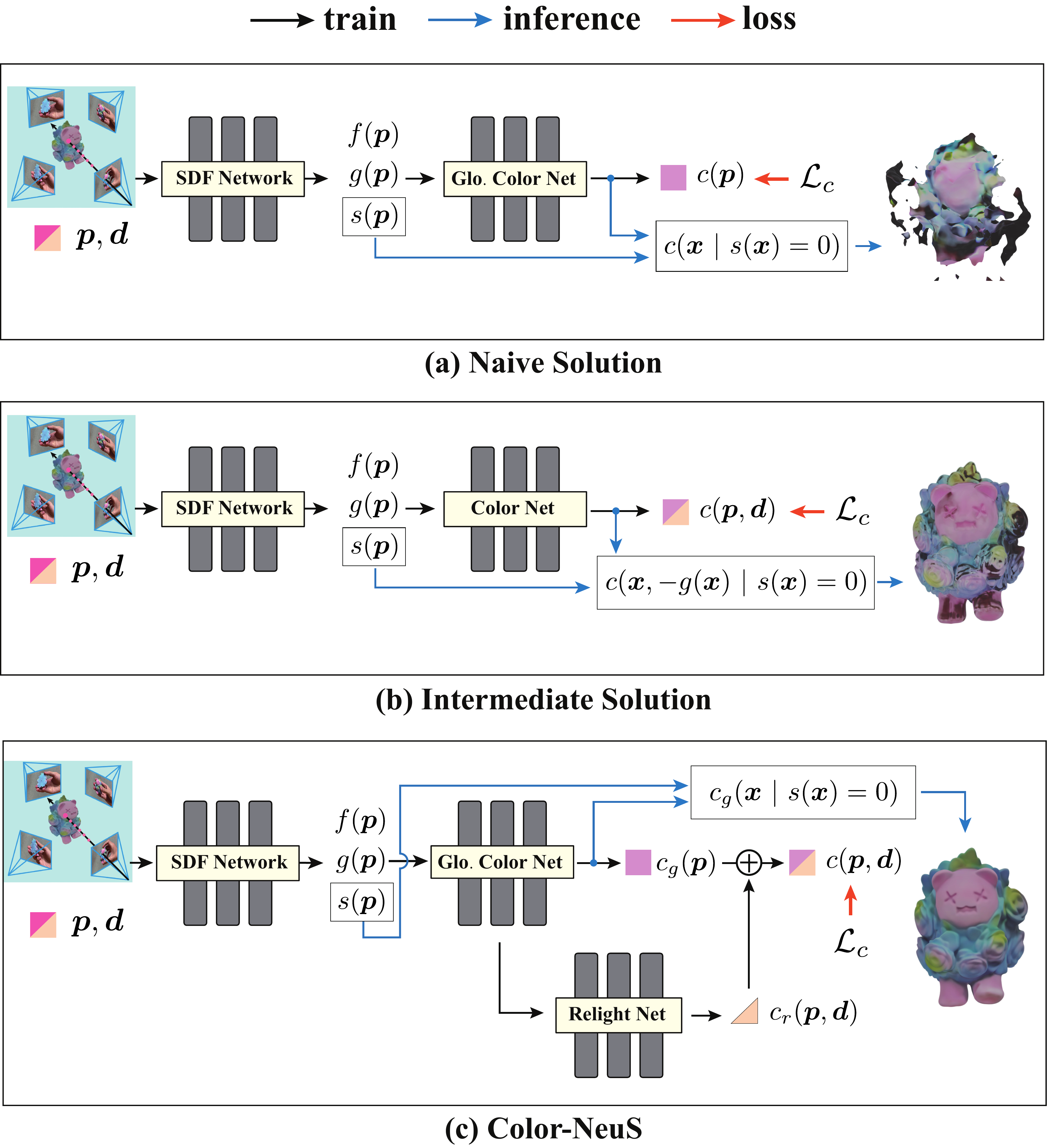}
    \end{center}
    \vspace{-15pt}
    \caption{\textbf{Illustration of the three solutions:} \textbf{(a) Naive Solution:} Utilize an SDF network to learn the implicit geometry, complemented with a global color network to grasp view-independent color; \textbf{(b) Intermediate Solution:} Employ a color network to learn view-dependent color and extract the surface color with a direction specified as per vertex normal; \textbf{(c) \method (Our Solution):} Make use of a global color network to learn view-independent color, and deploy a relight network to compensate for variations with the view direction. During inference, only the global view-independent color will be used.}
    \label{fig:pipeline}
\end{figure}

\subsection{Intermediate Solution}
\label{sec:intermediate}
The intermediate (interm.) solution is to utilize a constant direction as an input to the color network in order to extract color information from the surface. For instance, in HHOR \cite{huang2022hhor}, the vertex color is defined based on the surface's normal direction. For any $x \in \mathcal{S}$, the normal direction is given by $-g(\boldsymbol{x})$. Consequently, the vertex color can be extracted using
\begin{equation}
    c(\boldsymbol{x})=\mathcal{M}(\boldsymbol{x}, -g(\boldsymbol{x}),f(\boldsymbol{x}), g(\boldsymbol{x})).
    \label{eq:naive_solu}
\end{equation}

Yet, given the considerable variation in an object's surface color under changing lighting conditions, it may be insufficient to assign a singular `direction color' as a representation of the object's true color. The inherent complexity associated with light-surface interactions calls for a more nuanced approach to color extraction.

\subsection{\method}

In our method \method, we propose a workflow to disassociate the global color from the view-dependent formulation, all the while preserving the appropriate geometry and obtaining a reasonable view-independent color for each vertex.
Specifically, we substitute the learning of a single view-dependent color component (as indicated in \cref{eq:color_net}) with the learning of a view-independent global color variable (\cref{sec:remove_view_dep}) coupled with a view-dependent relighting effect (\cref{sec:relight}).

\subsubsection{Removing View-Dependence}\label{sec:remove_view_dep}

Our proposed solution begins with the removal of the view-dependent term from \cref{eq:color_net}, which transitions the model to a global color network, as:
\begin{equation}
    c_g(\boldsymbol{p})=\mathcal{M}_g(\boldsymbol{p}, f(\boldsymbol{p}), g(\boldsymbol{p})).
    \label{eq:color_net_rmd}
\end{equation}
This formulation matches the naive solution presented in \cref{eq:naive}, an ideal starting point for extracting the view-independent vertex color.
However,
solely employing this equation (ignoring the view-dependence) for optimizing network to convergence will violate the \cref{property:neus}.
Consequently, this contravention can prevent the density \emath{\sigma(\bm{p})} from condensing onto the object surface, inadvertently affecting the learned geometry.

Therefore, our proposed solution (\method) is to separate the inference procedure from optimizing (training).
During optimizing, we re-integrate the global color with a residual term that incorperates the view direction.
The outcome of this integration meets the view requirement set forth in the volume rendering formulation \cref{eq:volume_render}, and preserves the \cref{property:neus} through the optimizing phase.

During inference, once the density \emath{\sigma(\bm{p})} has condensed to the object surface, the shape of object (\emath{s(\bm{p}) = 0}) can be guaranteed. Additionally, \emath{c_g(\boldsymbol{p})} acquires the capability to express the vertex color.
Consequently, we can extract the vertex color on the surface (where $x \in \mathcal{S}$, that is \emath{s(\bm{x}) = 0}) during inference using:
\begin{equation}
    c_g(\boldsymbol{x})=\mathcal{M}_g(\boldsymbol{x}, f(\boldsymbol{x}), g(\boldsymbol{x})).
    \label{eq:extract_color_global}
\end{equation}

\subsubsection{Coupling Relighting Effect}\label{sec:relight}

To maintain the model's performance after removing view-dependence in \emath{c_g(\boldsymbol{x})}, we introduce a relighting network to compensate for the discarded view-dependent term. The relighting network functions with respect to the position, direction, and view-independent color, generating a small view-dependent color adjustment for each point. This can be expressed as:
\begin{equation}
    c_r(\boldsymbol{p}, \boldsymbol{d})=\mathcal{R}_g(c_g(\boldsymbol{p}), \boldsymbol{p}, \boldsymbol{d}, g(\boldsymbol{p})).
    \label{eq:relight_net}
\end{equation}
Subsequently, the ultimate color of each point is computed by integrating the relighting effect with the global color, as:
\begin{equation}
    c(\boldsymbol{p}, \boldsymbol{d}) = \Psi(\Psi^{-1}(c_g(\boldsymbol{p})) + c_r(\boldsymbol{p}, \boldsymbol{d})),
    \label{eq:relighted}
\end{equation}
where $\Psi$ and $\Psi^{-1}$ represent the sigmoid and inverse sigmoid function, respectively.
At present, the color is a fusion of a global color (view-independent) and a relighting effect (view-dependent).
The $c(\boldsymbol{p}, \boldsymbol{d})$ is then incorporated in \cref{eq:volume_render} to compute the volume integral $C$.

\subsection{Optimization}

The rendered color $C$ along the sampled rays can be computed using \cref{eq:volume_render,eq:sigma,eq:weights}. Let $\widehat{C}$ denote the ground truth color, we can define the color loss as:
\begin{equation}
    \mathcal{L}_c =\frac{1}{N_r}\sum_{i=1}^{N_r} \| \widehat{C}_i - C_i \|_2^2 ,
    \label{eq:color_loss}
\end{equation}
where $N_r$ denotes the number of sampled rays.

To enforce the optimized neural representation fulfills a valid SDF, we impose the eikonal regularization \cite{eikonal} on the SDF prediction, as:
\begin{equation}
    \mathcal{L}_e = \frac{1}{N_rN_p}\sum_{i,j}^{N_r,N_p}(\left \| \nabla s(\boldsymbol{p}_{i,j})\right \|_2 - 1)^2 ,
    \label{eq:eikonal_loss}
\end{equation}
where $N_p$ is the number of sampling points on each ray.

In a bid to draw the global color closer to the actual color, we impose a constraint on the mean value of the relight color $c_r$ to be zero, as:
\begin{equation}
    \mathcal{L}_r = \frac{1}{N_rN_p}\sum_{i,j=1}^{N_r,N_p} {c_r}_{i,j}(\boldsymbol{p}, \boldsymbol{d}).
    \label{eq:relight_loss}
\end{equation}
This strategy prompts the global color network to learn colors under average lighting conditions.
In other words, this minimizes the impact of the relighting network on the global color.
This loss term is necessary because we cannot directly supervise global color for the reasons mentioned in naive solution \ref{sec:naive}.

In the context of object reconstruction, an intuitive method involves using object (foreground) segmentation to eliminate background elements.
In scenarios where foreground segmentation is available and free of object-scene occlusion, we recommend incorporating a mask loss $L_m$
\begin{equation}
    \mathcal{L}_m = \frac{1}{N_r} \sum_{N_r} BCE(M, \hat{O})~,
    \label{eq:mask_loss}
\end{equation}
where $\hat{O}=\sum_j w_j$ is the cumulative weights along a camera ray,
and $M$ is a binary mask that signifies if a ray is within the boundaries of object segmentation.

In summary, our training loss is calculated as per \cref{eq:overall_loss}, where $\lambda_c$, $\lambda_e$, $\lambda_r$, and $\lambda_m$ are hyperparameters.
\begin{equation}
    \setlength{\abovedisplayskip}{3pt}
    \setlength{\belowdisplayskip}{3pt}
    \mathcal{L} = \lambda_c\mathcal{L}_c + \lambda_e\mathcal{L}_e + \lambda_r\mathcal{L}_r + \lambda_m\mathcal{L}_m .
    \label{eq:overall_loss}
\end{equation}

In addition to optimizing the network parameters mentioned above, for our wild dataset we also optimize the camera pose following NeRF$--$ \cite{wang2021nerfmm}. As in GNeRF \cite{meng2021gnerf}, we utilize a continuous 6D vector \( \bm{r} \in \mathbb{R}^6 \) to represent 3D rotations, which has been demonstrated to be more suitable for learning \cite{Zhou_2019_6dpose}. Jointly optimizing formulation can be expressed as:
\begin{equation}
    \setlength{\abovedisplayskip}{3pt}
    \setlength{\belowdisplayskip}{3pt}
    \Theta^*,\Pi^* = \argmin_{\Theta,\Pi} \mathcal{L}(\Theta,\Pi) ~,
\end{equation}
where $\Theta$ and $\Pi$ refer to the network parameters and camera pose, respectively.

\section{Experiment}
\label{sec:experiment}

\subsection{Implementation Details}
We random sample 1024 rays per batch with 8 random images and train our model for 100k iterations on a single NVIDIA A10 GPU. Following NeuS, the learning rate is first linearly warmed up from $0$ to $5 \times 10^{-4}$ in the first 5k iterations, and then controlled by the cosine decay schedule to the minimum learning rate of $2.5 \times 10^{-5}$. For all datasets, we set $\lambda_c$ to $1.0$, $\lambda_e$ to $0.1$, $\lambda_r$ to $1.0$.
For datasets where no object segmentation is available (OmniObject3D) or where  object-scene occlusion occurs (IHO-Video), we set $\lambda_m$ to $0.0$.
In contrast, for other datasets, we set $\lambda_m$ is $0.1$.
If a segmentation mask is available, we apply a sample strategy so that the amount of light falling inside the mask gradually increases from 50\% to 80\% during training.

\begin{figure}[H]
    \begin{center}
        \includegraphics[width=0.85\linewidth]{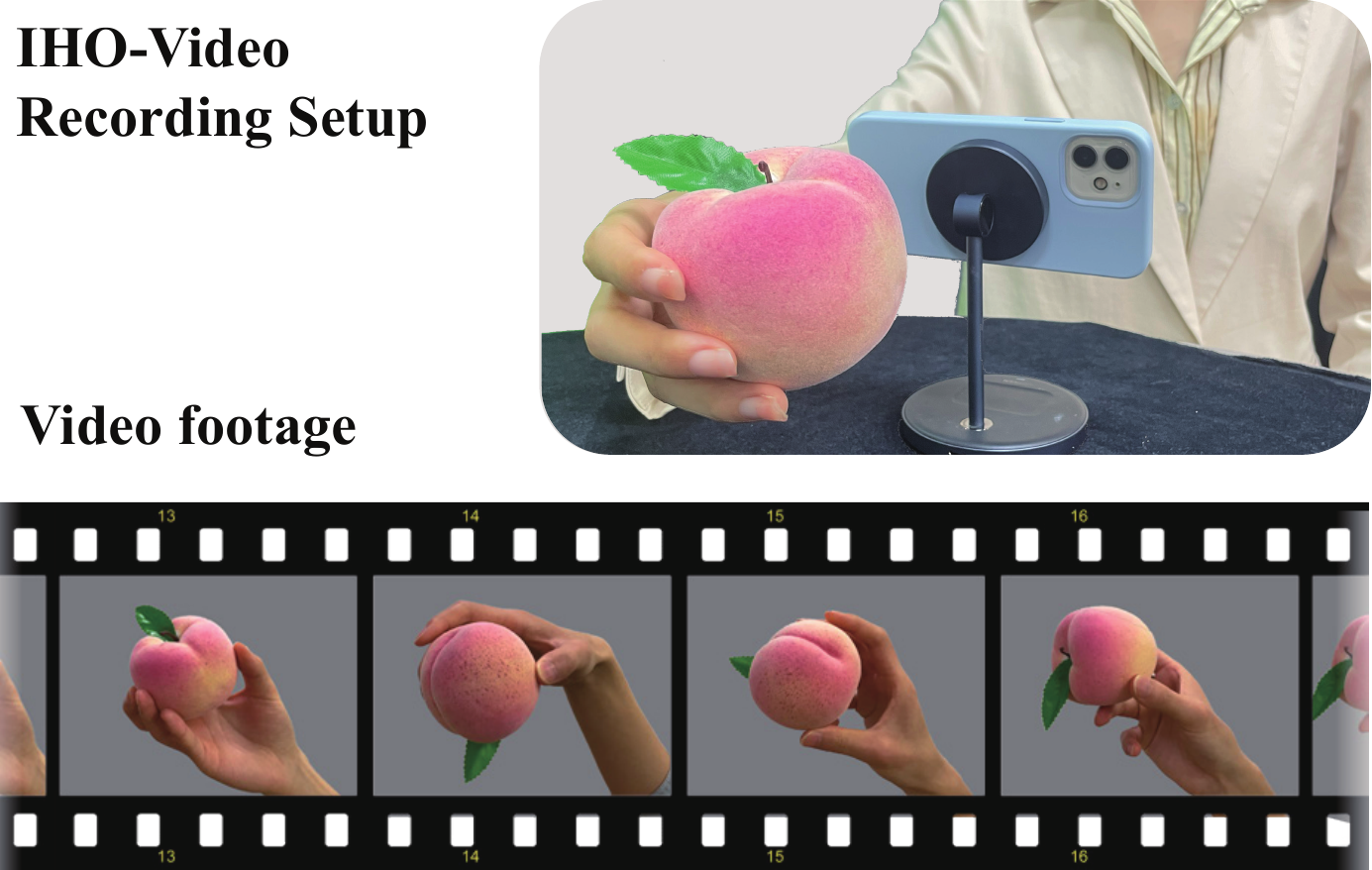}
        \caption{
            \textbf{Illustration of the monocular recording setup of the IHO-Video, HOS task.}
            The footage was captured using a uncalibrated mobile phone camera (default 1080p-30fps).
            This simple setup allows the quick deployment of our method,
            enabling users to collect data on their objects of interest with ease.
        }
        \label{fig:hos_video_footage}
    \end{center}
\end{figure}

\begin{figure}[H]
    \begin{center}
        \includegraphics[width=0.85\linewidth]{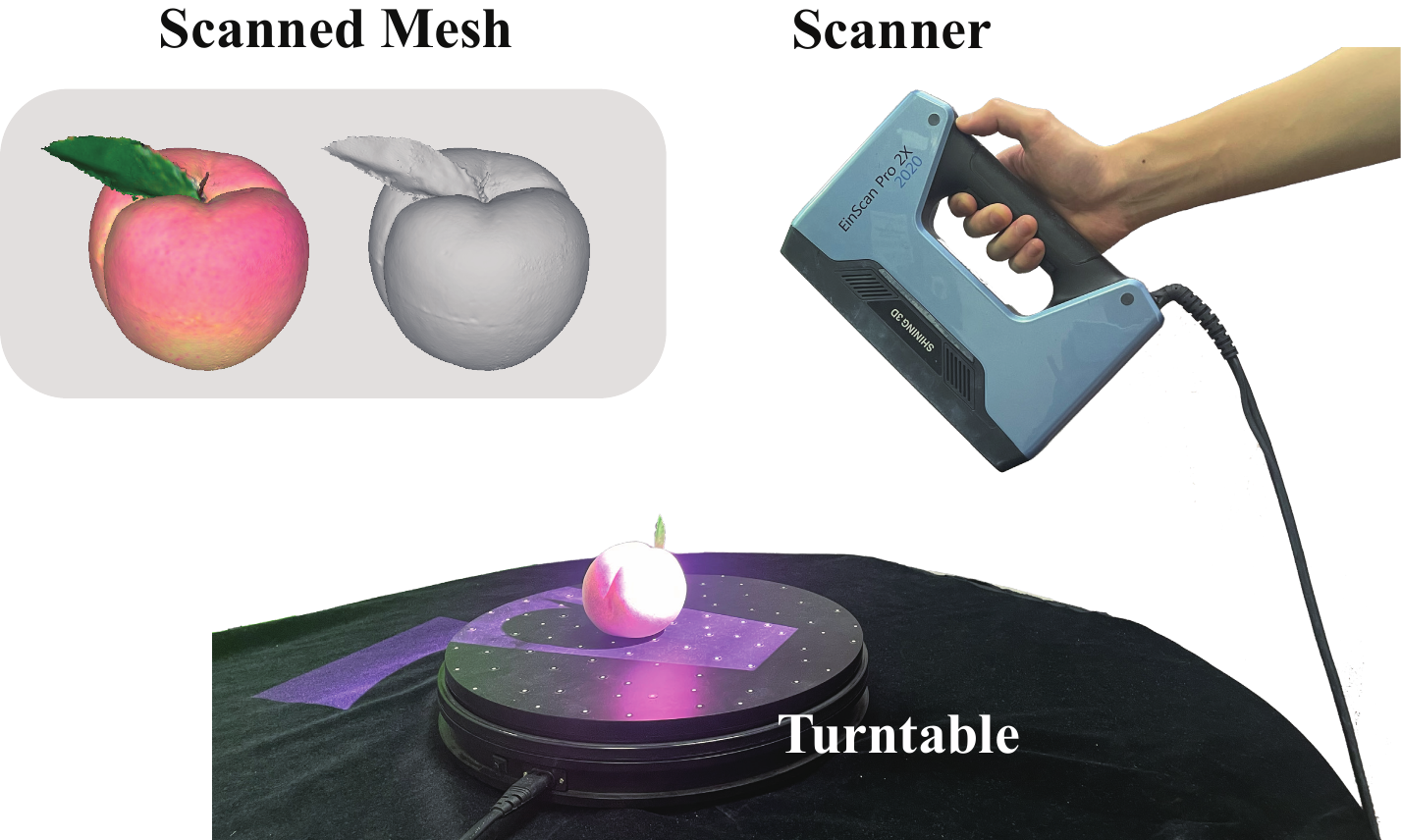}
        \caption{\textbf{Illustration of hand-held scanning system based on Structure-Light 3D scanner.} To fully scan the object, the subject must separately scan and then align its top and bottom surfaces. The scanned meshes serve as the ground-truth in the HOS task.}
        \label{fig:scan_table}
    \end{center}
\end{figure}
\vspace{-23pt}

\begin{figure*}[!t]

    \begin{center}
        \includegraphics[width=\linewidth]{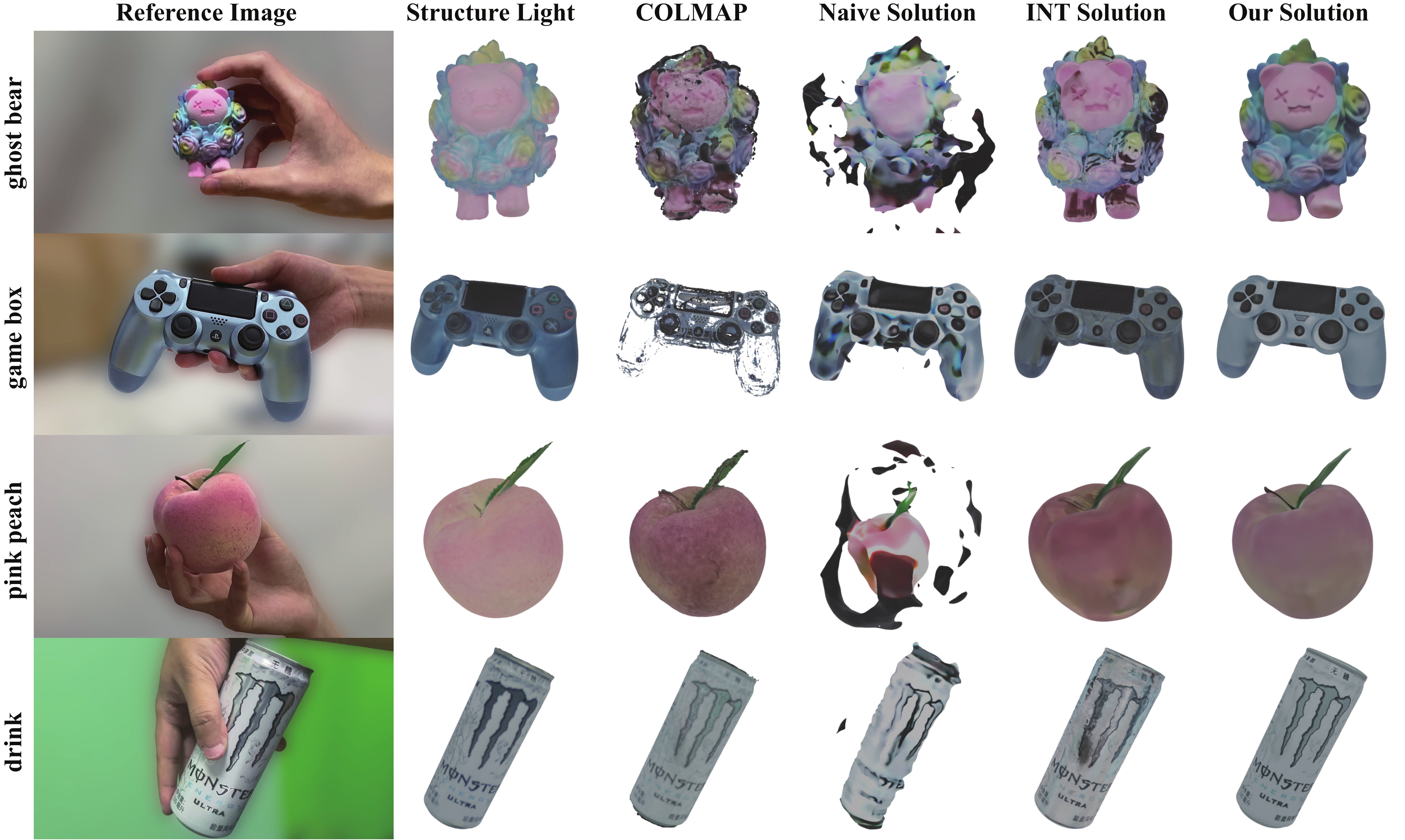}
        \vspace{-15pt}
        \caption{Results on IHO Video.}
        \label{fig:exp_iho}
    \end{center}

    \begin{center}
        \includegraphics[width=\linewidth]{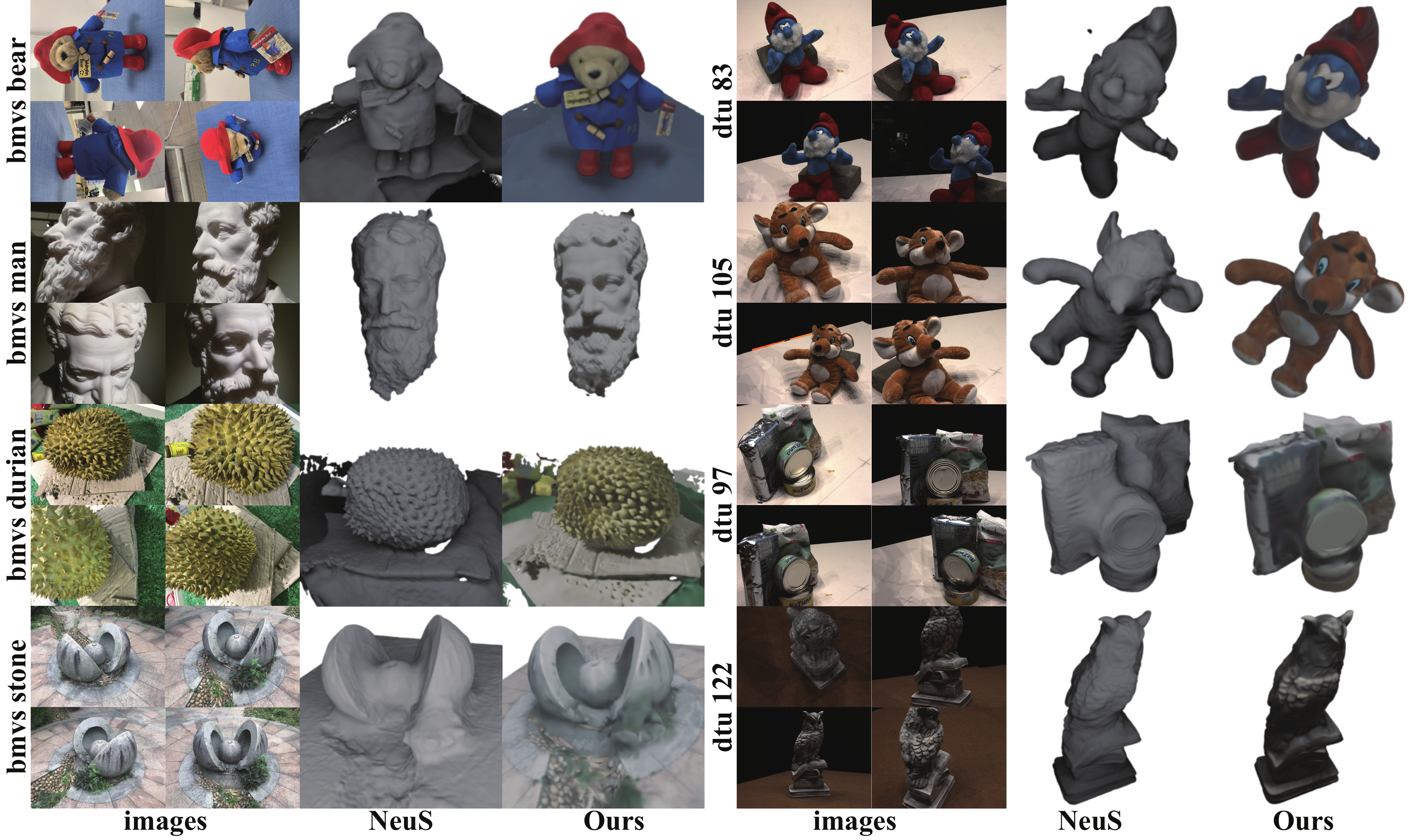}
        \vspace{-15pt}
        \caption{Results on BlendedMVS dataset and DTU dataset.}
        \label{fig:public}
    \end{center}
    \vspace{-10pt}

\end{figure*}

\subsection{Empirical Evaluation - Hand-held Object Scan}
We initiate the evaluation by deploying \method to a real-world challenging task: `hand-held object scanning' (HOS).
In HOS, to reveal the object appearance from all directions, the hand must continuously rotate and flip the object, thereby causing continual variation in occlusion and shadow effects.
Our performance evaluation is conducted in comparison with \method's alternative solutions.
To evaluate these solutions both quantitatively and qualitatively,
we have curated an original validation set termed `In-Hand Object Video' (IHO-Video).

\qheading{IHO-Video Dataset}
This dataset includes four video sequences, each documenting a hand-held object under a monocular recording setup (as illustrated in \cref{fig:hos_video_footage}). The footage was captured using an iPhone with resolution set to 1920$\times$1080.
Notably, these videos are characterized by their distinct dynamic attribute where objects and hands move in coordination, resulting in significant instances of dynamic occlusion and shadow interference.
Furthermore, surface reflections that fluctuate as the objects move add an additional layer of complexity.
To extract object masks for each frame, we employ Track-Anything \cite{yang2023track}, a method grounded in Segment-Anything \cite{kirillov2023segany} for segmentation and XMem \cite{cheng2022xmem} for instance tracking.
To obtain camera poses, we utilize the off-the-shelf Structure-from-Motion toolkit COLMAP \cite{schoenberger2016sfm, schoenberger2016mvs}.
Additionally, we used a professional hand-held Structured Light 3D scanner\footnote{\href{https://www.einscan.com/handheld-3d-scanner/einscan-pro-2x-2020/}{SHINING 3D EinScan Pro 2X 2020}} to obtain the ground-truth 3D object mesh, as shown in \cref{fig:scan_table}.

\begin{figure}[!t]
    \begin{center}
        \includegraphics[width=\linewidth]{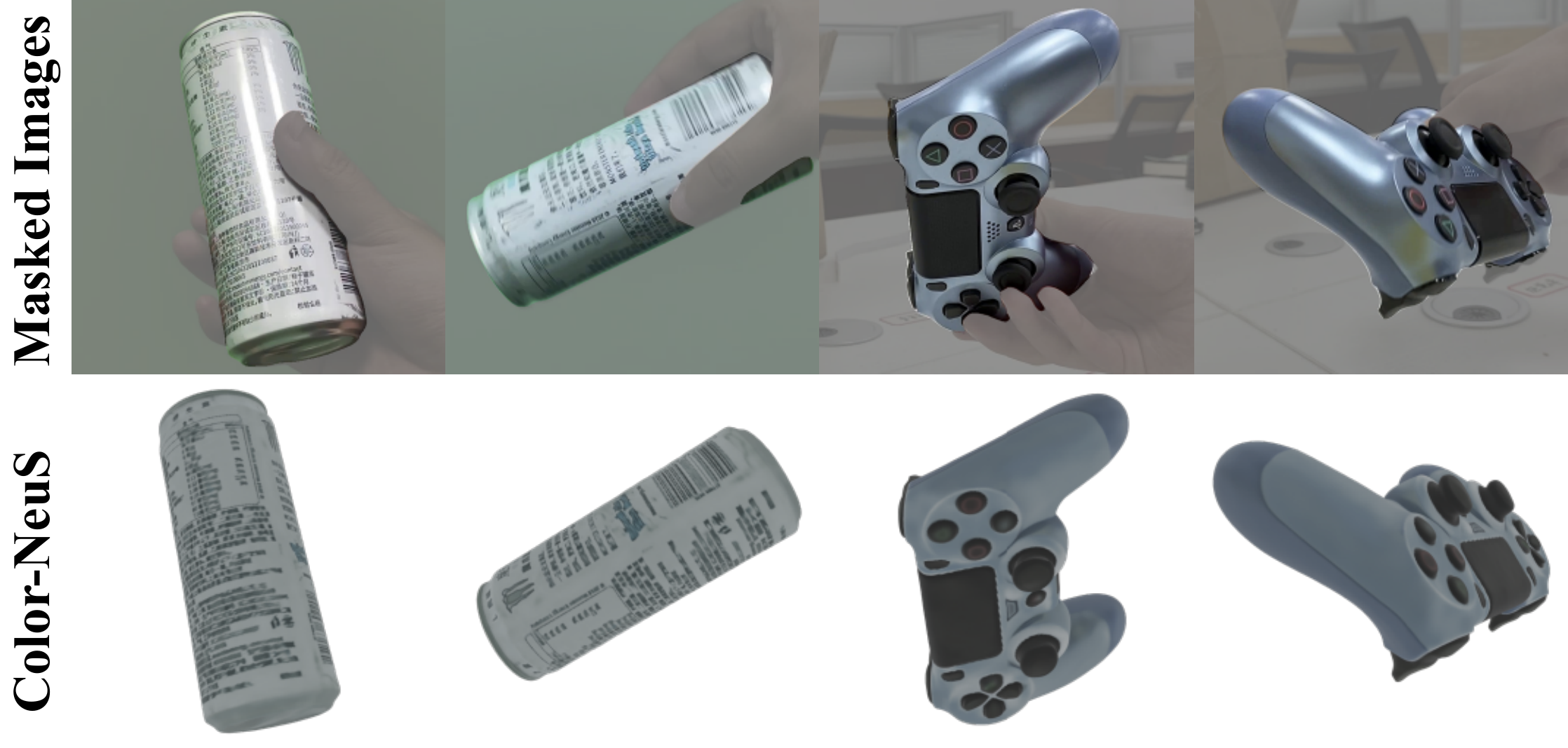}
        \caption{\textbf{Examples of occlusion and reflection phenomena in our IHO-Video dataset.}
            The first row showcase the images used for training. The gray area represent the background elements that are eliminated by the mask.
            The second row showcase the output meshes in the image-aligned view.
            As illustrated, \method consistently demonstrates robust performance in reconstructing both the shape and texture of the objects.
        }\vspace{-10pt}
        \label{fig:occlu_reflect}
    \end{center}
\end{figure}

\begin{figure}[!t]
    \begin{center}
        \includegraphics[width=\linewidth]{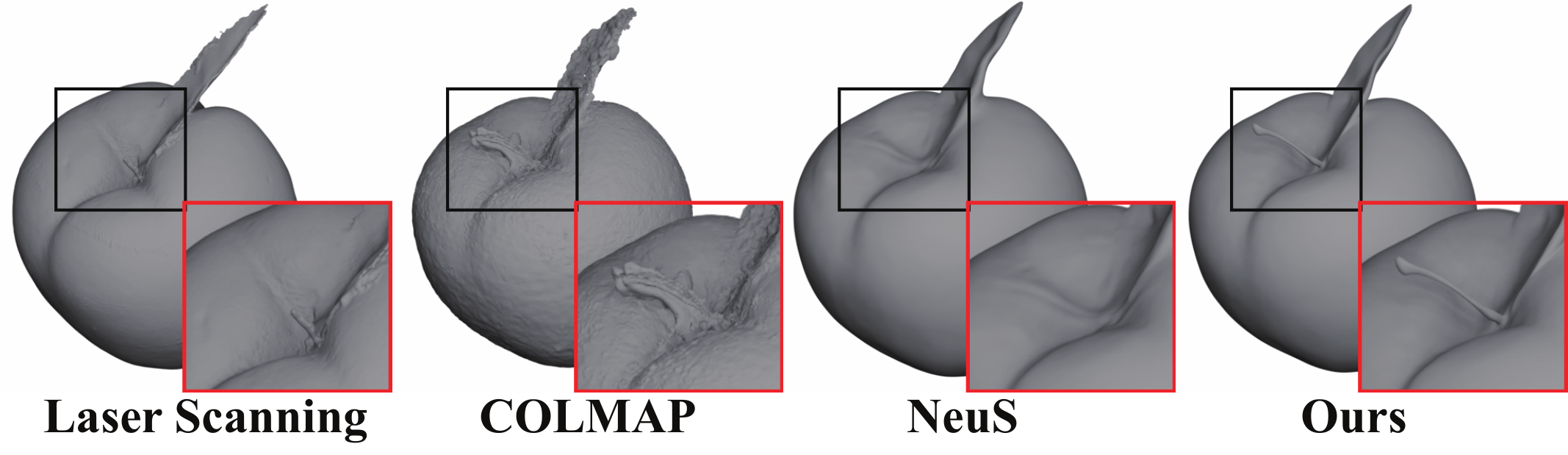}
    \end{center}
    \vspace{-10 pt}
    \caption{This illustration demonstrates that, among the four methods we applied to the pink peach sequence, only our approach successfully reconstructs the correct geometry of the peach rod.}
    \label{fig:mesh_pinkpeach}
    \vspace{-10pt}
\end{figure}
\qheading{HOS Task Evaluation}
We compared the results of \method with those derived from structured light scanning (see \cref{fig:scan_table}) and COLMAP's dense reconstruction.
Additionally, our solution was compared with both the naive and intermediate solutions discussed in \cref{sec:naive,sec:intermediate}. The respective results are depicted in \cref{fig:exp_iho}.

When scanning with structured light, the additional fill-in light results in color renderings divergent from those observed in the video footage.
The results of the Poisson reconstruction stemming from COLMAP are also not satisfactory due to the imperfect camera poses and low-resolution point cloud representation.
Besides, the naive solution was unable to accurately learn the object's geometry.
The interm. solution, despite successfully learning the correct geometry, faced difficulties dealing with occlusion and reflection, resulting artifacts such as dark spots appearing on the object mesh.
Contrarily, \method adeptly managed such disturbances.
\cref{fig:occlu_reflect} showcase several examples of occlusion and reflection.
In spite of these challenges, \method consistently exhibit robust performance in modeling the shape and appearances of the object surfaces.
Furthermore, its geometric reconstruction results aligned well with the ground truth (see \cref{tab:iho_cd}), showcasing its efficacy.

Our method has demonstrated considerable enhancement in object geometry learning. As evident in \cref{fig:mesh_pinkpeach}, traditional methods such as structured-light scanning, COLMAP, and NeuS fall short in accurately learning the structure of the peach rod. In contrast, our proposed \method exhibits commendable proficiency. This superior performance is attributed to \method's ability to discern view-independent color, preventing the misinterpretation of the pod as a black section on the surface of the peach body.

\subsection{ Evaluation on Public Datasets}
\vspace{-5pt}
\label{sec:exp_public}
\qheading{OmniObject3D. \cite{wu2023omniobject3d}}
OmniObject3D is a comprehensive 3D object dataset characterized by an extensive vocabulary and a large number of high-quality, real-scanned 3D objects. It includes 6,000 scanned objects drawn from 190 everyday categories. Furthermore, it also provides object-centric, photorealistic multi-view images rendered using Blender \cite{blender}, thus facilitating a wide range of experiments and analyses. Due to this dataset's extensive size, in this paper, we show the effect on two subsets from the rendered images, namely, the `doll' set and the `toy\_animals' set.
\begin{figure}[!t]
    \begin{center}
        \includegraphics[width=\linewidth]{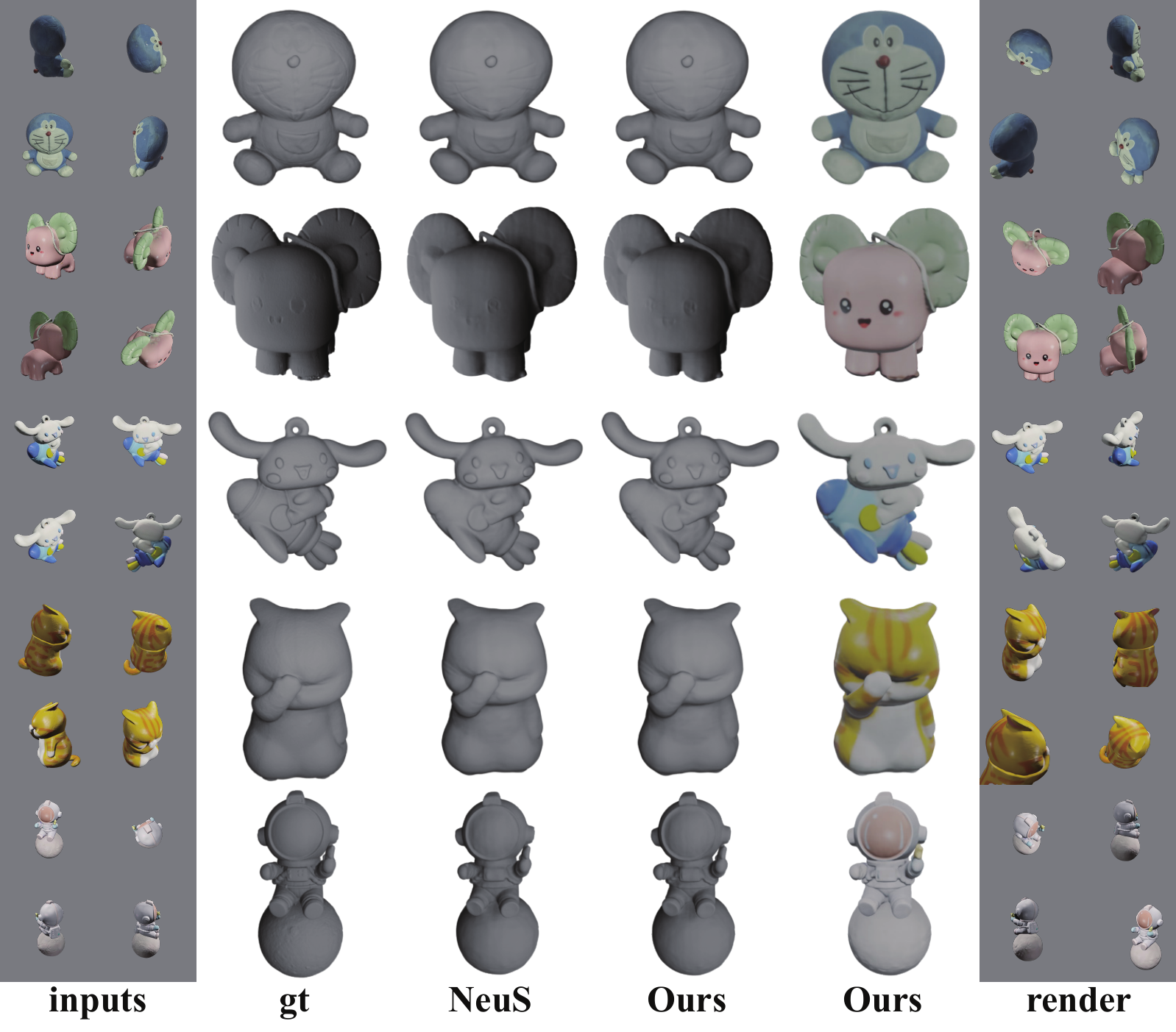}
    \end{center}
    \vspace{-15pt}
    \caption{Results on OmniObject3D dataset.}
    \label{fig:exp_omni}
    \vspace{-10pt}
\end{figure}

\qheading{DTU. \cite{jensen2014large}}
The DTU dataset comprises a broad diversity of materials, appearances, and geometric structures, posing significant challenges for reconstruction algorithms, including handling non-Lambertian surfaces and delicate, thin structures. Consistent with the NeuS \cite{wang2021neus} method, we employ 15 scenes from this dataset for our experiments.

\qheading{BlendedMVS. \cite{yao2020blendedmvs}}
Same with NeuS \cite{wang2021neus}, we also conducted tests on 8 challenging scenes drawn from the low-resolution subset of the BlendedMVS dataset. In this dataset, each image exhibits a resolution of $768 \times 576$ pixels, accompanied by corresponding masks.

\qheading{HOD. \cite{huang2022hhor}} Hand-held Object Dataset (HOD) is a dataset contains 35 objects, which is divided into two subsets named Sculptures and Daily Objects. Because our task is different with HHOR (they reconstruct hand and object in a tightly held pose), we only use the `Giuliano' sequence in Sculptures subset for evaluation.
\eofparagraph\bigskip

\vspace{-5pt}
\noindent We compare \method with original NeuS \cite{wang2021neus} on BlendedMVS \cite{yao2020blendedmvs} and DTU \cite{jensen2014large} dataset as shown in \cref{fig:public}, \method can extract surface color while maintain the ability to learn geometry. In \cref{fig:exp_omni}, we evaluate \method on OmniObject3D \cite{wu2023omniobject3d} dataset. Beyond the ability to learn geometry, due to the design of religinting network, \method can also maintain the ability for novel view synthesis. Checkout our supplementary materials for more results.

\begin{figure}[!t]
    \begin{center}
        \includegraphics[width=\linewidth]{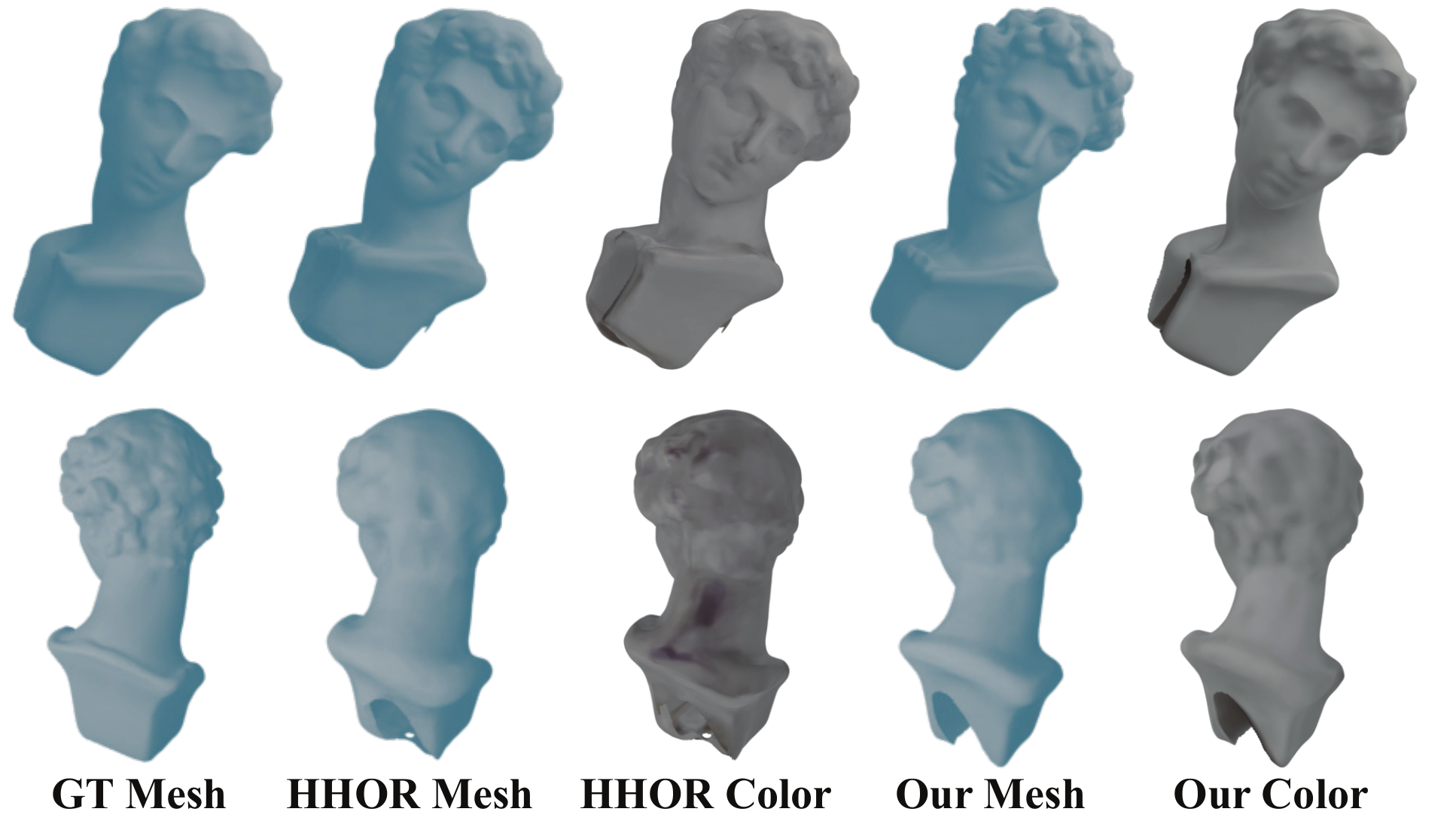}
    \end{center}
    \vspace{-15pt}
    \caption{Results on Giuliano sequence in HOD dataset.}
    \label{fig:hhor}
    \vspace{-15pt}
\end{figure}

Our approach was also contrasted with that of HHOR \cite{huang2022hhor} on the HOD \cite{huang2022hhor} dataset. HHOR uses the intermediate solution for color extraction. As evidenced in \cref{fig:hhor} and \cref{tab:hhor_cd}, \method demonstrates the ability to reconstruct the surface color with fewer impurities and enhanced geometrical accuracy.

\subsection{Quantitative Results}
\vspace{-5pt}
\label{sec:metrics}
In this section, we employ the Chamfer Distance on OmniObject3D \cite{wu2023omniobject3d}, Giuliano (a sequence of HOD \cite{huang2022hhor}), and IHO-Video as a quantitative measure to evaluate the efficacy of our method. It is pertinent to mention that the geometric aspect of our approach is founded on NeuS \cite{wang2021neus}. We will demonstrate that the geometric learning capability of our method is at least on par with NeuS. In other words, our modifications will not degrade the performance inherent to NeuS. From a theoretical standpoint, the application of our technique to enhancements of NeuS, such as PET-NeuS \cite{wang2023petneus}, should yield superior geometric outcomes.

\begin{table}[ht]
    \centering
    \begin{small}
        \begin{tabular}{lccccc}
            \shline
            doll ID & 002           & 008           & 049           & 074           & 085           \\
            \hline
            NeuS    & 0.35          & 2.31          & 0.55          & 0.42          & 0.97          \\
            Ours    & \textbf{0.27} & \textbf{1.91} & \textbf{0.53} & \textbf{0.40} & \textbf{0.74} \\
            \shline
        \end{tabular}
    \end{small}
    \vspace{-5pt}
    \caption{Chamfer Distance on OmniObject3D.}
    \label{tab:omni_cd}
    \vspace{-20pt}
\end{table}

\begin{table}[H]
    \centering
    \begin{small}
        \begin{tabular}{l|cc}
            \shline
            Giuliano         & HHOR & Ours          \\
            \hline
            Chamfer Distance & 8.25 & \textbf{5.89} \\
            \shline
        \end{tabular}
    \end{small}
    \vspace{-5pt}
    \caption{Chamfer Distance on Giuliano sequence.}
    \label{tab:hhor_cd}
    \vspace{-10pt}
\end{table}
\begin{table}[ht]
    \centering
    \begin{small}
        \begin{tabular}{l|cccc}
            \shline
            IHO Video  & COLMAP        & Naive  & NeuS          & Our           \\
            \hline
            ghost bear & 7.84          & 197.65 & \textbf{1.18} & 1.32          \\
            game box   & 3.54          & 12.57  & 0.78          & \textbf{0.55} \\
            pink peach & 12.69         & 450.20 & 19.39         & \textbf{8.13} \\
            drink      & \textbf{6.68} & 60.52  & 11.29         & 8.60          \\
            \shline
        \end{tabular}
    \end{small}
    \vspace{-5pt}
    \caption{Chamfer Distance on IHO Video.}
    \label{tab:iho_cd}
    \vspace{-10pt}
\end{table}
We first normalize each mesh to unit size ($1m$), then utilize the point-to-point iterative closest point (ICP) algorithm to align the reconstructed mesh with the ground truth.
In all the tables, the Chamfer Distance (CD) is defined based on squared distance, with a unit of $1 cm^{2}$. Results is shown in \cref{tab:omni_cd,tab:hhor_cd,tab:iho_cd}.

\subsection{Ablation Study}
\label{sec:ablation}

\begin{figure}[H]
    \begin{center}
        \includegraphics[width=\linewidth]{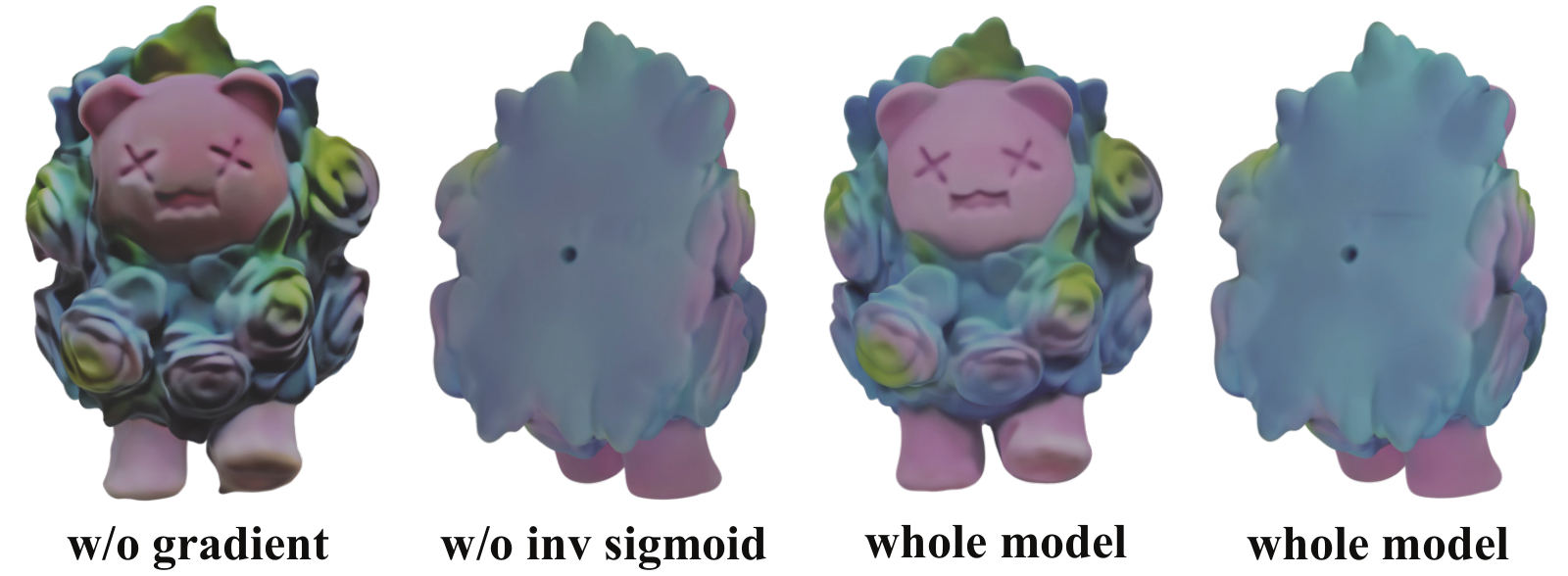}
    \end{center}
    \vspace{-15pt}
    \caption{Ablation study.}
    \label{fig:ablation}
    \vspace{-10pt}
\end{figure}

In the relighting network, the gradient of the Signed Distance Function (SDF), denoted as $g(\boldsymbol{p})$, is incorporated as input. This decision's efficacy is demonstrated in an ablation study, the results of which can be viewed in \cref{fig:ablation}. We believe that this gradient input offers accurate geometric information to the network, consequently leading to a faithful color distribution.

In another ablation study, we investigated the significance of using the inverse sigmoid function. In \cref{eq:relighted}, we initially apply the inverse sigmoid function, denoted by $\Psi^{-1}$, to the RGB value. Subsequently, we introduce the relight color, followed by a sigmoid function, $\Psi$. An alternative approach would be to directly restrict the relighted color to the interval [0, 1]:
\vspace{-4pt}
\begin{equation}
    c(\boldsymbol{p}, \boldsymbol{d}) = \mathrm{clamp}(c_g(\boldsymbol{p})+\Psi(c_r(\boldsymbol{p}, \boldsymbol{d}))-0.5,0,1).
    \label{eq:relighted_clamp}
    \vspace{-3pt}
\end{equation}

The outcomes are displayed in \cref{fig:ablation}. The findings suggest that refraining from employing the inverse sigmoid function tends to result in marginally darker color outputs.

\section{Conclusion}
\label{sec:conclusion}

In this paper, we introduce \method, a novel approach to 3D implicit textured surface reconstruction which is compatible with any NeuS-like models. By isolating the view-dependent element from the neural radiance field and employing a relighting network to sustain volume rendering, \method is able to effectively retrieve surface color while accurately reconstructing the surface detail. We put \method the test on a demanding in-hand object scanning task using our personally collected sequences as well as several public datasets. The results demonstrate \method's capability to reconstruct neural implicit surfaces with accurate color representation.

\noindent\rule[0.3ex]{\linewidth}{1.0pt}
{\small
    \textbf{Acknowledgments} This work was supported by the National Key R\&D Program of China (No. 2021ZD0110704), Shanghai Municipal Science and Technology Major Project (2021SHZDZX0102), Shanghai Qi Zhi Institute, and Shanghai Science and Technology Commission (21511101200).
}
{
    \small
    \bibliographystyle{configs/ieeenat_fullname}
    \bibliography{egbib}
}
\clearpage
\setcounter{page}{1}
\maketitlesupplementary

\section{Additional details}
\label{sec:additional_details}
\begin{figure}[H]
    \begin{center}
        \includegraphics[width=0.9\linewidth]{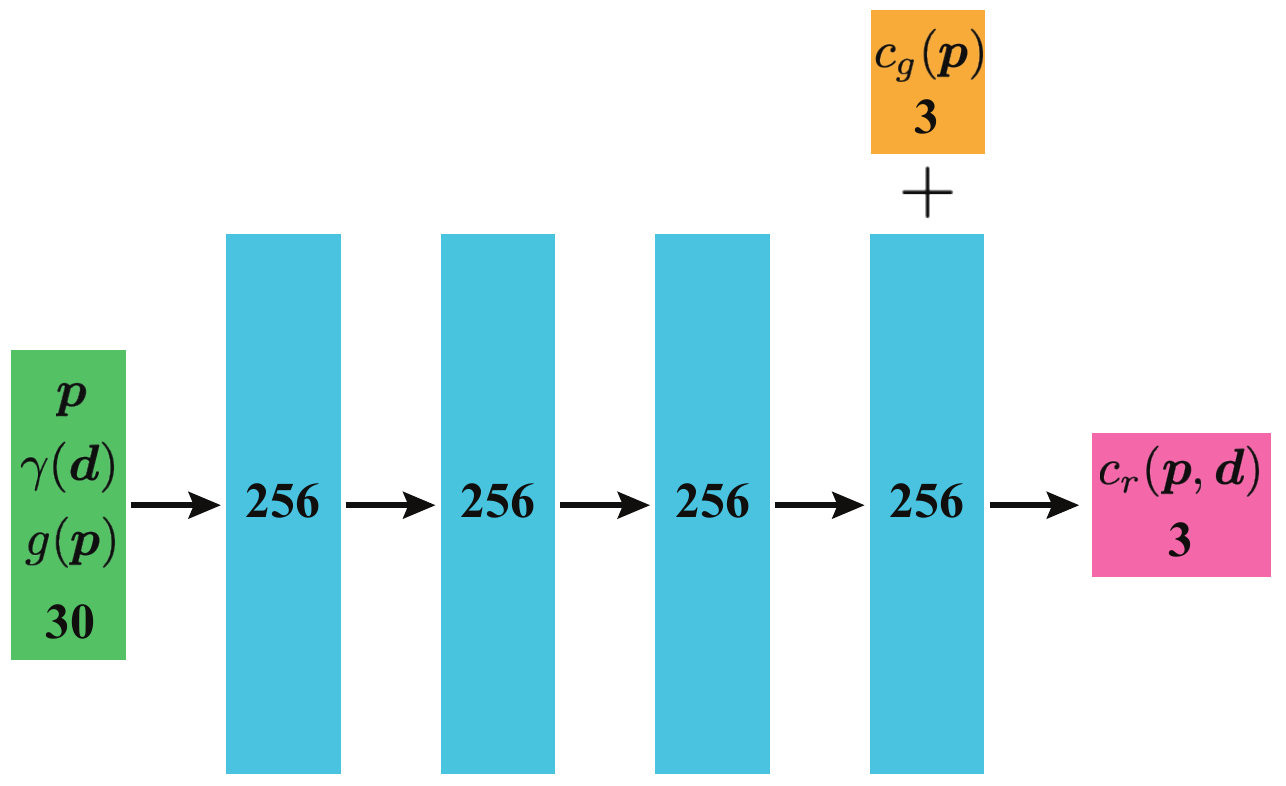}
    \end{center}
    \vspace{-15pt}
    \caption{Relight network architecture.}
    \label{fig:relightnetwork}
\end{figure}

In NeRF \cite{mildenhall2020nerf}, positional encoding, denoted by \(\gamma(\cdot)\), is utilized to allow the network to capture high-frequency details. In \method, positional encoding is applied to the spatial location \(\boldsymbol{p}\) with 6 frequencies within the SDF network, and to the view direction \(\boldsymbol{d}\) with 4 frequencies in the relight network. We adopt a similar architecture for the SDF network and global color network as found in NeuS \cite{wang2021neus}. The architecture of the relight network, comprising 4 hidden layers each with a hidden size of 256, is depicted in \cref{fig:relightnetwork}. The global color information is fed into the final layer.

\section{More Results}
\label{sec:more_results}
\begin{figure}[!t]
    \begin{center}
        \includegraphics[width=\linewidth]{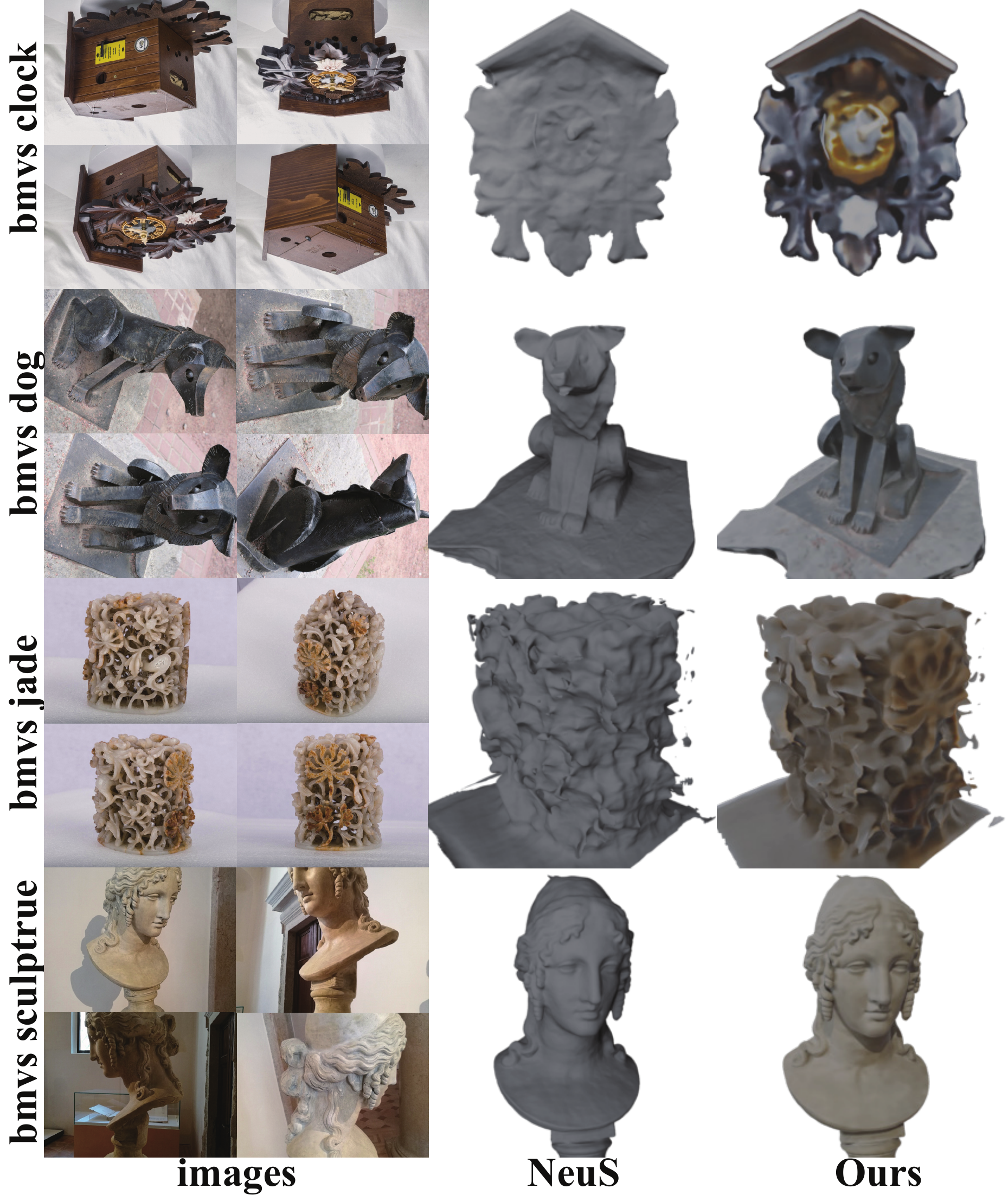}
    \end{center}
    \vspace{-15pt}
    \caption{More results on the BlenderMVS dataset.}
    \label{fig:bmvs_more}
\end{figure}

\begin{figure}[!t]
    \begin{center}
        \includegraphics[width=\linewidth]{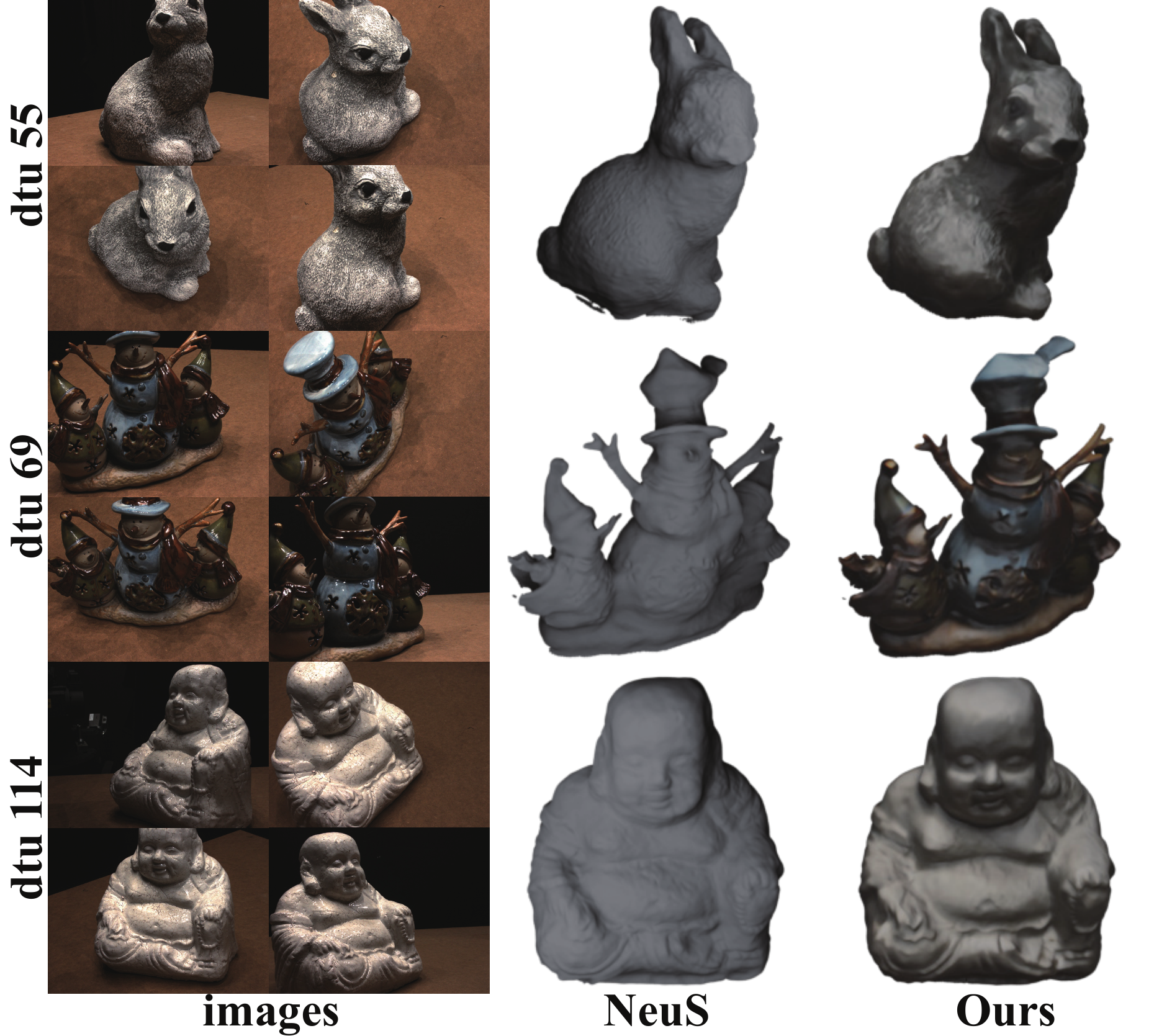}
    \end{center}
    \vspace{-15pt}
    \caption{More results on the DTU dataset.}
    \label{fig:dtu_more}
    \vspace{-5pt}
\end{figure}

In this section, we present additional results on public datasets. Specifically, the quantitative results related to OmniObject3D are shown in \cref{tab:omni_cd_more}, while the qualitative outcomes for both BlenderMVS and DTU are depicted in \cref{fig:bmvs_more} and \cref{fig:dtu_more}, respectively.

\begin{table}[ht]
    \centering
    \begin{small}
        \begin{tabular}{lccccc}
            \shline
            toy\_animals ID & 001           & 005           & 016           & 019           & 059           \\
            \hline
            NeuS            & \textbf{0.53} & 12.05         & 4.36          & 3.43          & \textbf{1.09} \\
            Ours            & 0.91          & \textbf{8.66} & \textbf{2.47} & \textbf{1.06} & 1.18          \\
            \shline
        \end{tabular}
    \end{small}
    \vspace{-5pt}
    \caption{More results of Chamfer Distance on OmniObject3D.}
    \label{tab:omni_cd_more}
    \vspace{-20pt}
\end{table}

\section{Rendering Quality}
\label{sec:rendering_quality}
\begin{figure*}[!t]

    \begin{center}
        \includegraphics[width=\linewidth]{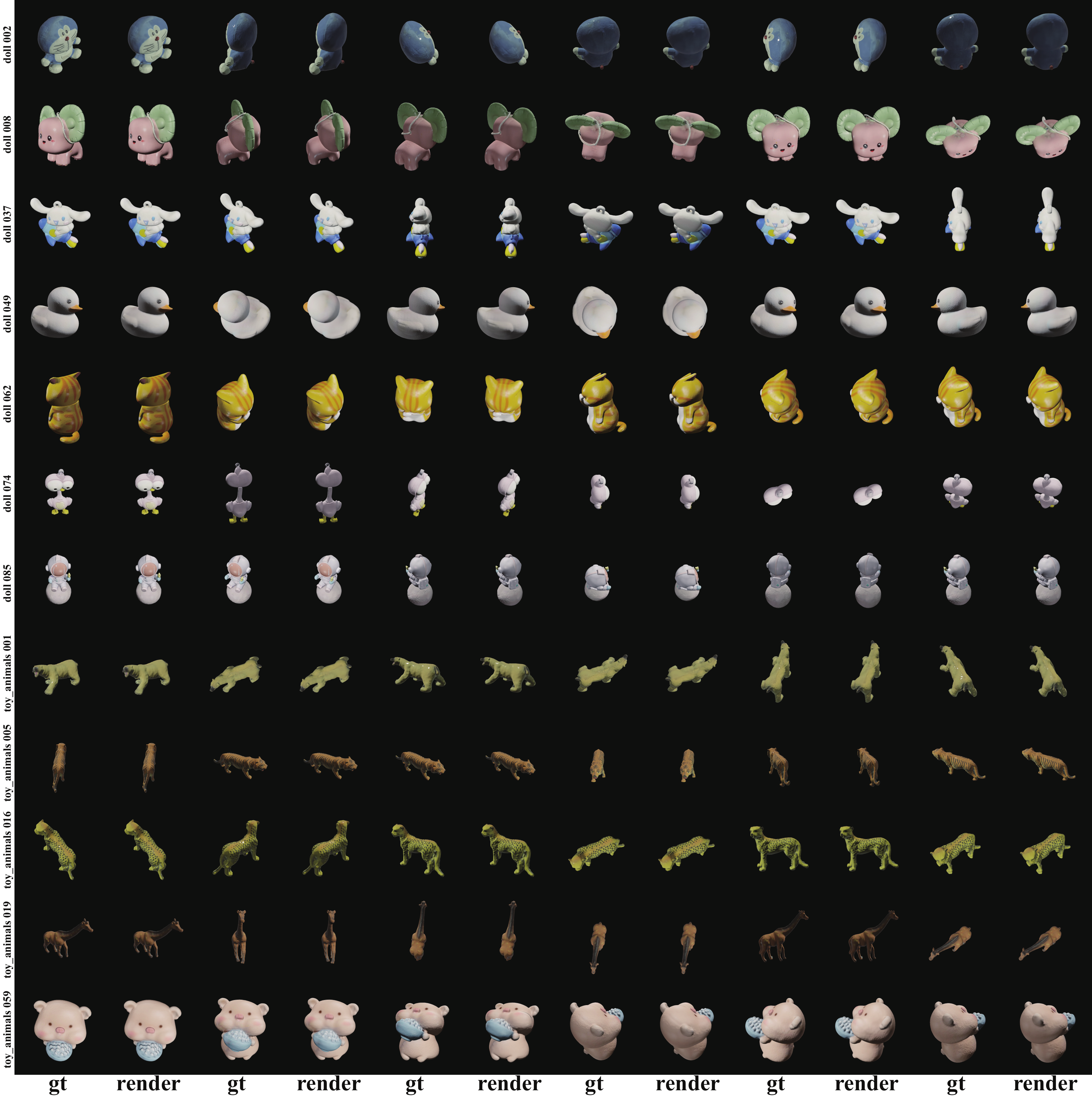}
        \vspace{-15pt}
        \caption{New view synthesis (2D render) results on the OmniObject3D dataset.}
        \label{fig:omni_render}
    \end{center}
    \vspace{-10pt}

\end{figure*}

To demonstrate that our method retains the capability of volume rendering and can perform new view synthesis, as described in \cref{sec:exp_public}, we conducted tests on the OmniObject3D dataset \cite{wu2023omniobject3d}. The results are presented in \cref{tab:omni_render} and \cref{fig:omni_render}. For each sequence, we allocated \(90\%\) of the images for training and the remaining \(10\%\) for testing. The training configurations were identical to those used in the main paper, and we ensured that our method, NeuS \cite{wang2021neus}, and NeRF \cite{mildenhall2020nerf} were trained under the same settings. The results reveal that our method can achieve rendering performance comparable to that of NeRF \cite{mildenhall2020nerf} and NeuS \cite{wang2021neus}.

\begin{table*}[!t]
    \centering
    \begin{small}
        \begin{tabular}{c|l|ccccccc|ccccc|c}
            \shline
            \multicolumn{2}{c|}{}               & \multicolumn{7}{c|}{doll ID} & \multicolumn{5}{c|}{toy\_animals ID} &                                                                                               \\
            \hline
            \multicolumn{2}{c|}{}               & 002                          & 008                                  & 037   & 049   & 062   & 074   & 085   & 001   & 005   & 016   & 019   & 059   & Mean          \\
            \shline
            \multirow{3}*{\rotatebox{90}{PSNR}} & NeRF                         & 37.45                                & 36.27 & 36.71 & 36.77 & 37.79 & 37.13 & 37.21 & 37.76 & 39.64 & 36.70 & 40.35 & 35.84 & 37.47 \\
                                                & NeuS                         & 37.87                                & 35.95 & 37.10 & 37.74 & 37.65 & 37.27 & 37.76 & 37.88 & 39.34 & 35.96 & 40.27 & 36.07 & 37.57 \\
                                                & Ours                         & 37.47                                & 35.20 & 36.67 & 37.43 & 37.43 & 37.23 & 37.28 & 37.82 & 39.01 & 35.69 & 39.55 & 35.66 & 37.20 \\
            \hline
            \multirow{3}*{\rotatebox{90}{SSIM}} & NeRF                         & 0.983                                & 0.975 & 0.987 & 0.979 & 0.979 & 0.987 & 0.985 & 0.985 & 0.987 & 0.980 & 0.989 & 0.982 & 0.983 \\
                                                & NeuS                         & 0.986                                & 0.975 & 0.989 & 0.983 & 0.979 & 0.989 & 0.988 & 0.986 & 0.987 & 0.978 & 0.990 & 0.984 & 0.984 \\
                                                & Ours                         & 0.985                                & 0.973 & 0.988 & 0.983 & 0.979 & 0.989 & 0.987 & 0.985 & 0.986 & 0.977 & 0.989 & 0.984 & 0.984 \\

            \shline
        \end{tabular}
    \end{small}
    \vspace{-5pt}
    \caption{Quantitative results for new view synthesis on the OmniObject3D dataset.}
    \label{tab:omni_render}
\end{table*}

\end{document}